\documentclass[12pt]{spieman}  
\usepackage{amsmath,amsfonts,amssymb}
\usepackage{graphicx}
\usepackage{setspace}
\usepackage{tocloft}

\makeatletter
\let\c@lofdepth\relax
\let\c@lotdepth\relax
\makeatother
\usepackage{subfigure}
\usepackage{color,xcolor}
\usepackage{multirow}
\usepackage{lineno}

\usepackage{algorithm}  
\usepackage{algpseudocode}  

\title{Pyramid Frequency Network with Spatial Attention Residual Refinement Module for Monocular Depth Estimation}

\author[a]{Zhengyang Lu}
\author[a,*]{Ying Chen}
\affil[a]{Key Laboratory of Advanced Process Control for Light Industry (Ministry of Education), Jiangnan University, Wuxi, China}

\cftpagenumbersoff{figure}
\cftpagenumbersoff{table} 
\begin{document} 
\maketitle


\begin{abstract}
Deep-learning-based approaches to depth estimation are rapidly advancing, offering superior performance over existing methods.
To estimate the depth in real-world scenarios, depth estimation models require the robustness of various noise environments.
In this work, a Pyramid Frequency Network(PFN) with Spatial Attention Residual Refinement Module(SARRM) is proposed to deal with the weak robustness of existing deep-learning methods.
To reconstruct depth maps with accurate details, the SARRM constructs a residual fusion method with an attention mechanism to refine the blur depth.
The frequency division strategy is designed, and the frequency pyramid network is developed to extract features from multiple frequency bands.
With the frequency strategy, PFN achieves better visual accuracy than state-of-the-art methods in both indoor and outdoor scenes on Make3D, KITTI depth, and NYUv2 datasets.
Additional experiments on the noisy NYUv2 dataset demonstrate that PFN is more reliable than existing deep-learning methods in high-noise scenes.
\end{abstract}


{\noindent \footnotesize\textbf{*}Ying Chen, \linkable{chenying@jiangnan.edu.cn} }

\begin{spacing}{2}   

\section{Introduction}

Although Deep Convolutional Neural Networks(DCNNs) have attracted great attention in recent decades and have shown significant progress in computer vision enabled by the availability of large-scale labelled datasets, these models lack the robustness of the human vision system.
It has been observed that numerous image corruptions, such as Gaussian noise, fog, overexposure, and blurring, can lead to tremendous degradation\cite{azulay2018deep}.
Improving network robustness is an important step towards safely deploying models in real-world scenarios.

Recently, the robustness of monocular depth estimation DCNNs has two common solutions, namely data augmentation and model redundancy.
Data augmentation, a technique used to increase the amount of data by adding slightly modified copies of existing data, is a widely-used and constantly applicable approach to train robust models. 
Models apply image transformations to training data, such as flipping, cropping, adding random noise and adding occlusions\cite{shorten2019survey}.
Data augmentation acts as a regularizer and helps reduce over-fitting when training DCNNs.
However, it is observed that data augmentation exacerbates the data imbalance and increases model training costs.
Simple data augmentation methods have inevitable limitations due to the overemphasizing features of homogeneous instances.
Model redundancy is another solution to enhance the robustness of model-based approaches\cite{nguyen2016redundancy}.
Though the model redundancy method increases model size, it allows the model to choose reliable features.
However, due to the notorious unreliability of DCNNs, model redundancy is an attractive option.

In this work, we propose a Pyramid Frequency Network(PFN) with Spatial Attention Residual Refinement Module(SARRM) for monocular depth estimation, which aims at handling the weak robustness of previous deep-learning methods.
Specifically, the key contributions of the {\bf PFN} can be summarized as follows:

\begin{itemize}
\item Our approach is inspired by the Fourier perspective on model robustness\cite{yin2019fourier}, and relies on frequency division and a feature pyramid network to extract frequency features.
The frequency pyramid is proposed to separate the image into different frequencies, which alleviates the periodic noise issue and guides the model to reconstruct accurate depths..
\item A spatial attention residual refinement module(SARRM) is presented for the information fusion between high-frequency and low-frequency domains.
As an auxiliary module, it not only helps extract features from the RGB domain, but also necessitates the combination of preliminary depth and detailed information recovered from multi-stage frequency bands.
Thus, it can improve the feature extraction capability of the model, and is also useful in cases where only high-noise or severe-corruption scene images are available.
\item We present experiments on three benchmark datasets, demonstrating that our {\bf PFN} significantly improves the performance of monocular depth estimation. In particular, on the KITTI depth dataset\cite{geiger2013vision}, the {\bf PFN} achieves a new state-of-the-art accuracy of 95.3\% when the threshold is $1.25$, whereas the previous best result was 95.1\%. 
In addition, extensive experiments show that our {\bf PFN} achieves better visual accuracy than most existing methods under noisy circumstances.

\end{itemize}

\section{Related work}

In the past few years, the DCNNs have been applied to many fields in computer vision with great success. We briefly survey techniques related to our work, namely model-based depth estimation.

Most deep-learning-based methods are fully supervised, requiring ground-truth depth during training. Depth estimation is fundamental for understanding the stereoscopic of scenes from 2D images. From the beginning of depth estimation research, the majority of work has concentrated on geometry-based algorithms\cite{flynn2016deepstereo}\cite{scharstein2002taxonomy} that rely on point correspondences between images and triangulation. 

A current popular approach is to construct a depth map using learning non-linear multi-channel to single-channel mapping, implemented as a CNN work\cite{wang2015towards}. Because the handcrafted features can only capture limited local information, the global relationship was incorporated using deep CNN and Markov Random Fields features\cite{saxena2008make3d}\cite{liu2014discrete}. 
Another novel approach to using global information was the Depth-transfer\cite{karsch2014depth}, which generates plausible depth maps from videos using non-parametric depth sampling.

Eigen\cite{eigen2014depth} employed two deep network stacks to incorporate global and local cues: one made a coarse global prediction based on the entire image, while the other refined this prediction locally. A single multi-scale convolutional network architecture\cite{eigen2015predicting} was applied for three independent computer vision tasks, namely depth prediction, normal surface estimation, and semantic labeling. 
Eigen\cite{eigen2015predicting} proposed another method that captured more image details, and refined predictions using a sequence of scales.

Furthermore, a deep convolutional neural field model\cite{liu2015learning} was presented for estimating depths from single monocular images, aiming to jointly explore the capacity of deep CNN and continuous Conditional Random Field(CRF). 
Following that, Liu proposed a deep structured learning scheme\cite{liu2015learning} for learning the unary and pairwise potentials of continuous CRF within a unified deep CNN framework.

Deep Ordinal Regression Network for Monocular Depth Estimation(DORN)\cite{fu2018deep} was proposed to discretize continuous depth into specified intervals and cast the depth network learning as an ordinal regression problem, and it showed how to involve ordinal regression in a dense prediction task via deep CNNs. 
Monocular Relative Depth Perception\cite{xian2018monocular} and Structure-guided Ranking Loss for depth prediction\cite{xian2020structure} also involved ordinal regression in the task of depth prediction.
Xian\cite{xian2018monocular} introduced a method to automatically generate dense relative depth annotations from web stereo images and proposed a ranking loss to deal with imbalanced ordinal relations, which enforces the network to focus on a set of hard pairs. 
Then, Xian\cite{xian2020structure} proved that the pair-wise ranking loss, combined with the structure-guided sampling strategies, benefits the depth prediction.

To address the issue of occluding contours, Ramamonjisoa introduced SharpNet\cite{ramamonjisoa2019sharpnet}, a method that predicts an accurate depth map given a single input color image, with particular attention to the reconstruction of occluding contours. 
To facilitate the guidance of densely encoded features to the desired depth prediction, Lee\cite{lee2019big} proposed a multi-scale local planar guidance technique for monocular depth estimation called Big to Small(BTS). This technique utilized novel local planar guidance layers located at multiple stages during the decoding phase.

Recently, Liu\cite{liu2017learning} proposed learning the affinity through a deep CNN with Spatial Propagation Networks, yielding better results comparing to the traditional manually designed affinity. However, depth refinement commonly necessitated a local affinity matrix rather than a global affinity matrix. 
To fuse the discrete ground-truth depth points measured by LIDAR sensors, Cheng\cite{cheng2018depth} presented the Convolutional Spatial Propagation Network(CSPN), in which the depths of all pixels are updated simultaneously within a local affinity.
For fast and accurate monocular depth estimation, Lu\cite{lu2021ga} proposed an effective generative adversarial network.
The approach demonstrated the feasibility of applying a dense-connected UNet for reducing information transmission loss and then fine-tuning the blur depth using the high-order Convolutional Spatial Propagation Network(GA-CSPN) with a modified discriminator loss function.
Additionally, the loss function of the discriminator was modified by adding the correlation loss, which is used to measure the similarity of real and fake labels.

Besides fully-supervised depth estimation, another alternative is to train depth estimation models using a corresponding view image as supervision. Godard et al.\cite{godard2017unsupervised} constructed Monodepth, an unsupervised Monocular Depth Estimation with Left-Right Consistency. Monodepth built a deep network to predict stereo pairs and predict their pixel disparities. 
Furthermore, Godard et al.\cite{godard2019digging} proposed the Monodepth2, a fully-resolution multi-scale sampling method that minimizes visual artifacts, and an auto-masking loss to ignore training pixels that violate camera motion assumptions.

For model-based depth estimation, the pyramid structure provided an efficient architecture to extract different types of information from various layers.
Liao\cite{liao2021adaptive} assumed that most of the depth hypotheses up-sampled from a well-estimated depth map were accurate. Based on this assumption, Liao introduced a pyramid multi-view stereo network based on adaptive depth estimation, which gradually refines and up-samples the depth map to the desired resolution. 
To enhance the robustness in ill-posed regions or new scenes, Tian\cite{tian2020depth} came up with a self-improving pyramid stereo network that can not obtain a direct regression disparity without complicated post-processing.

Lee\cite{lee2018single} proposed a single-image depth estimation method based on the Fourier domain analysis(FDA). The first step is generating multiple depth map candidates by cropping images with various ratios and then combining the multiple candidates in the frequency domain.

\subsection{Preliminaries}

Although great progresses has been made in the field of monocular depth estimation, estimating an accurate dense depth from a single image remains challenging, even for humans. 
One of the reasons is that it is an intractable problem due to the complexity and high noise of application scenarios in the real world.
Therefore, to estimate the depth of real-world scenarios, these models require the robustness of various high-noise environments.
For example, it has been observed that commonly occurring image corruptions, such as Gaussian noise, fog, overexposure, contrast change, and blurring, can lead to significant performance degradation\cite{dodge2017study}.

Inspired by the Fourier perspective on model robustness\cite{yin2019fourier}, we apply Fourier domain processing for depth estimation tasks.
In contrast to the RGB domain, the Fourier domain makes it possible for the separation of most noises from corrupted images.
However, Fourier features, which are complex frequency domain features, are notoriously difficult to apply directly to feature fusion tasks.
In the depth estimation task, frequency division is a feasible option that divides the Fourier features into multi-stage frequency bands that each contain different components of depth information.

\begin{figure*}[htbp]
\centerline{\includegraphics[width=1\linewidth]{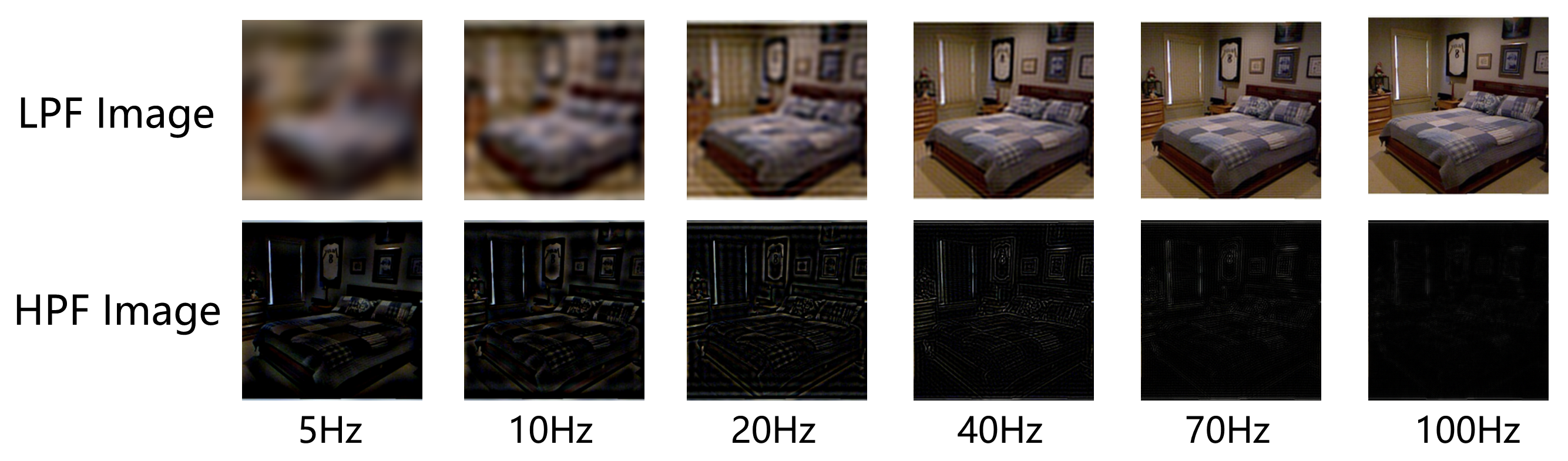}}
\caption{Images that have been processed by the LPF and HPF retain various degrees of valid information.}
\label{Res4filter}
\end{figure*}

To obtain the frequency response, each depth map is estimated from source images that have been divided into multiple frequency bands.
The pipeline involves first separating RGB images into each frequency band that needs to be evaluated for the percentage of valid information.
As a result of frequency division, Fig.\ref{Res4filter} displays the images that have been separated into multiple frequency bands.
From Fig.\ref{Res4filter}, the images which pass through the High Pass Filter(HPF) retain the texture and edge information.
Theoretically, the outputs of the Low Pass Filter(LPF) preserve aperiodic signals which contain the most efficacious information for depth estimation.

\begin{figure*}[htbp]
\centerline{\includegraphics[width=0.7\linewidth]{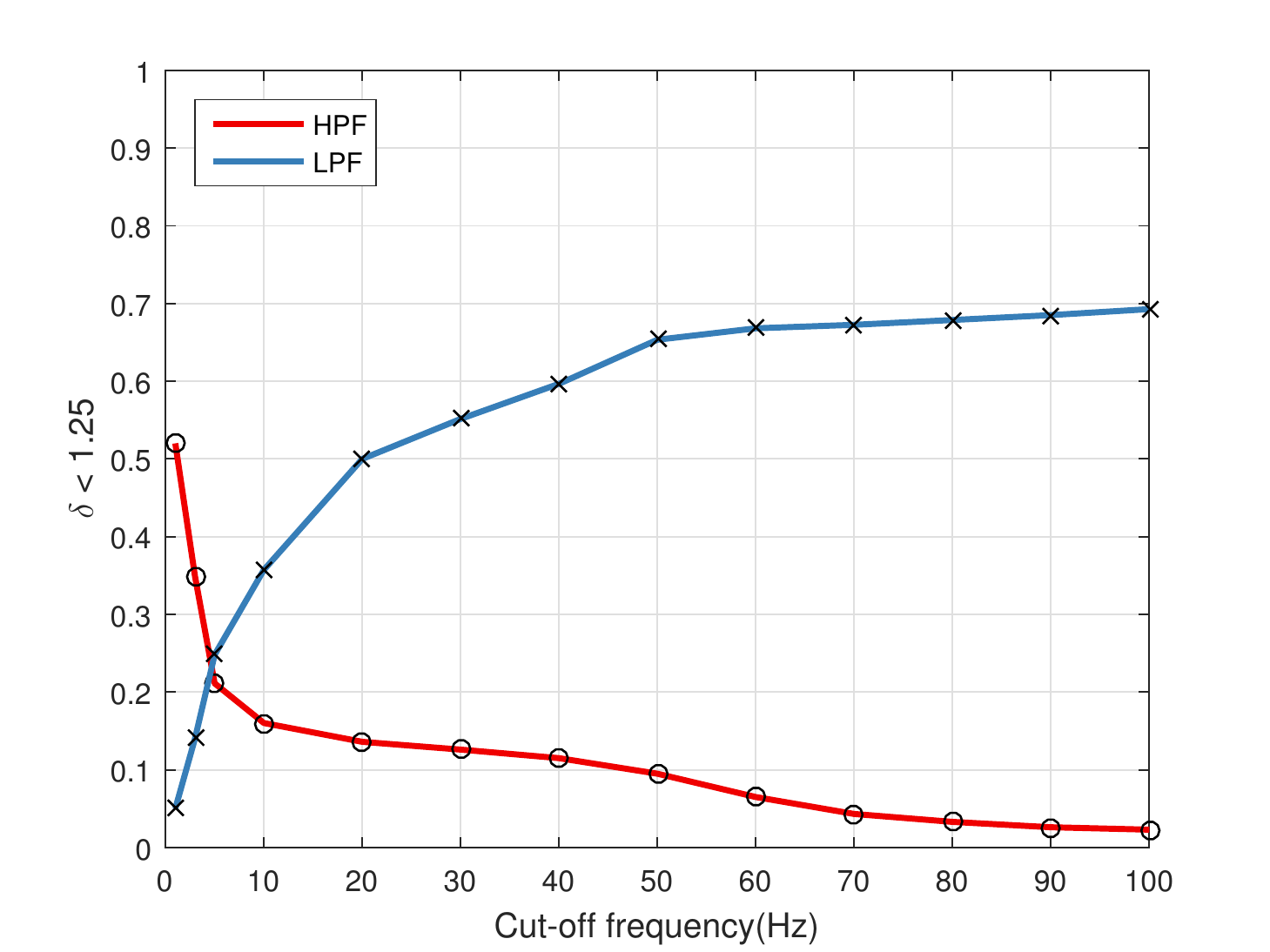}}
\caption{Images that passed through the LPF and HPF preserve various levels of valid information.}
\label{Acc4freqprob}
\end{figure*}

The absolute relative errors of the estimated depths obtained from multiple frequency bands are shown in Fig.\ref{Acc4freqprob}.
It can be observed from Fig.\ref{Acc4freqprob} that the basic U-net backbone\cite{ronneberger2015u} can achieve high accuracy, which is evaluated by the percentage of depth where the ratio of estimated and ground-truth depth is less than a threshold $\delta$, using information from the input that would be unrecognizable to humans.
From another perspective, the accuracy curve represents the percentage of valid information. 
However, the image pass through HPF loses most of valuable depth information and the most of depth information exists in the low-frequency domain.
The valid information percentage of each frequency band is proportional to the absolute value of the fitted curve gradient.
Therefore, the crucial information for the depth estimation task is mainly preserved in the frequency band between 5Hz and 30Hz.


\section{Method}

\begin{figure*}[ht]
\centerline{\includegraphics[width=0.8\linewidth]{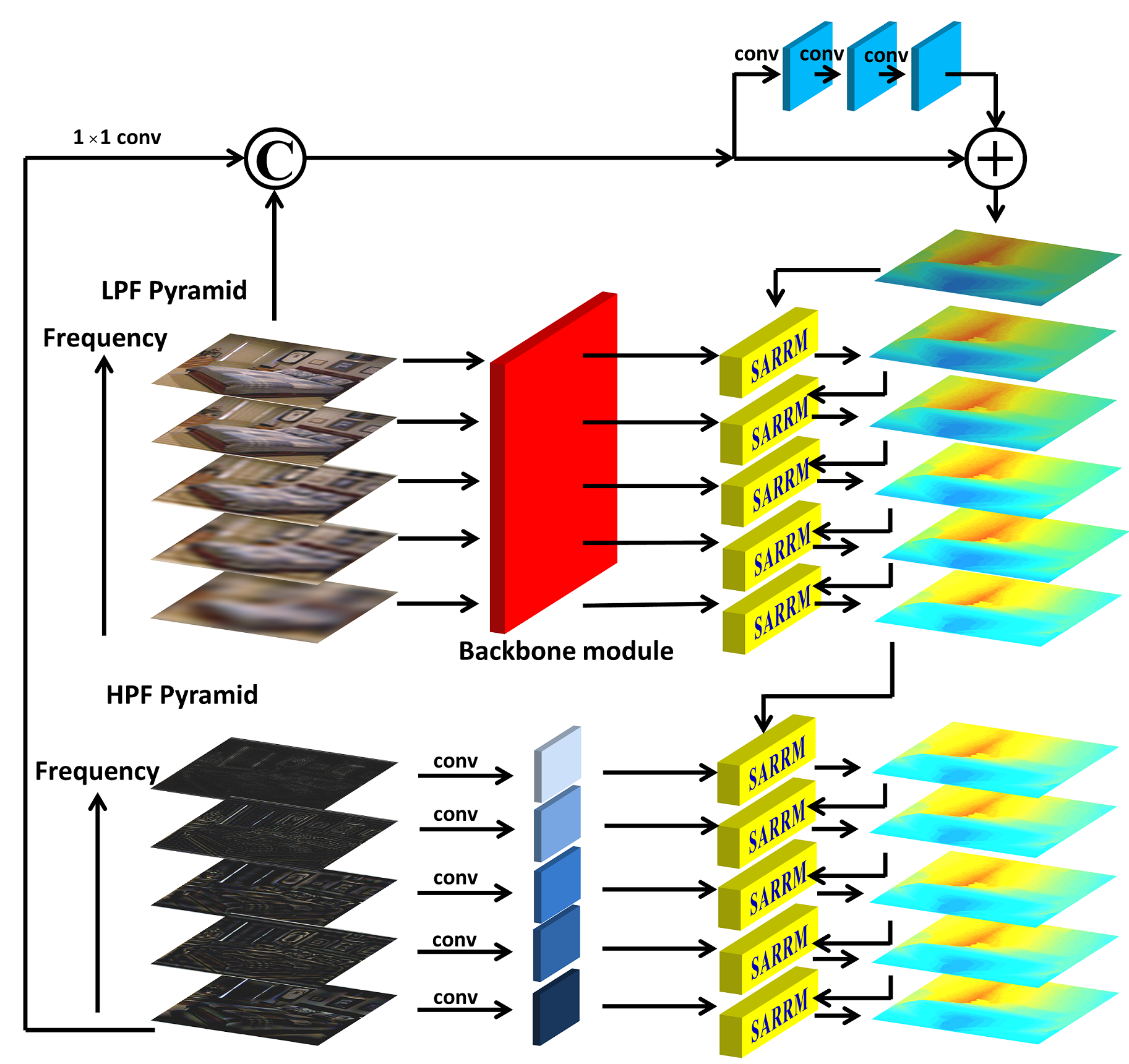}}
\caption{Pyramid Frequency Network architecture.}
\label{PFN}
\end{figure*}

In order to fuse the Fourier information from multiple frequency bands, we construct a fully-supervised DCNN based on the frequency division called Pyramid Frequency Network({\bf PFN}) with Spatial Attention Residual Refinement Module(SARRM). 
To improve the network's ability to adapt multi-layer input, we present a pyramid architecture to boost model fusion and help it take advantage of the deeper backbone.
We then describe a simple and effective fusion method namely Spatial Attention Residual Refinement Module(SARRM), and demonstrate its superiority over previous fusion methods.
Fig.\ref{PFN} overviews the proposed {\bf PFN}.

Through frequency division, the information from multi-stage frequency bands can be obtained, thereby increasing the details of the reconstruction depth and improving the accuracy.

The backbone network, shared by the preliminary depth estimation modules, extracts features from each frequency pyramid stage.
In our experiments, we adopt a High-Resolution network(HR-net)\cite{sun2019deep} as the preliminary depth estimation backbone.
In the high-frequency module, each HPF pyramid layer, followed by a convolution layer, refines the blurred depth map with high-level information such as texture and edge features.
The SARRM module, which combines the residual network with the spatial attention channel, helps to fuse the blur depth and single-layer features.

\subsection{Pyramid frequency network}

The 2D image frequency division is implemented by the 2D Fast Fourier Transform(2D-FFT) method which is an effective implement of 2D Fourier Transform(2D-FT).

Fig.\ref{PFN} shows an overall architecture of {\bf PFN} where the left part is the Pyramid frequency network.
As it can be seen from Fig.\ref{PFN}, the 2D image is divided into subdivisions on multi-stage frequency bands in order. 
Compared to the FDA\cite{lee2018single}, which first applied the frequency domain method in depth estimation, our work separates images directly by multiple frequency bands, instead of by unreasonably various ratios of image cropping.
By analyzing the correlation of Fourier components in Fig.\ref{Acc4freqprob}, we can observe that the low-frequency subdivisions contain more depth information and for the LPF images provide more basic information, such as rough depth plane.
Different from the LPF pyramid that separates the global depth planar from the original image, the HPF pyramid provides local texture depth information that guides features to the high resolution with clear boundaries.
For the order of subdivisions on multi-stage frequency bands, it follows the increasing order of information.

Frequency division can help to fuse the network features from multi-stage frequency bands, which is beneficial for the positive periodic signal. 
For the negative periodic signal, such as noise, the network tends to ignore the irrelevant information during the feature fusion.
The bandwidth and cut-off frequencies in the frequency domain division method will be determined in ablation experiments.

\subsubsection{End-to-end backbone}

\begin{figure*}[htbp]
\centerline{\includegraphics[width=1\linewidth]{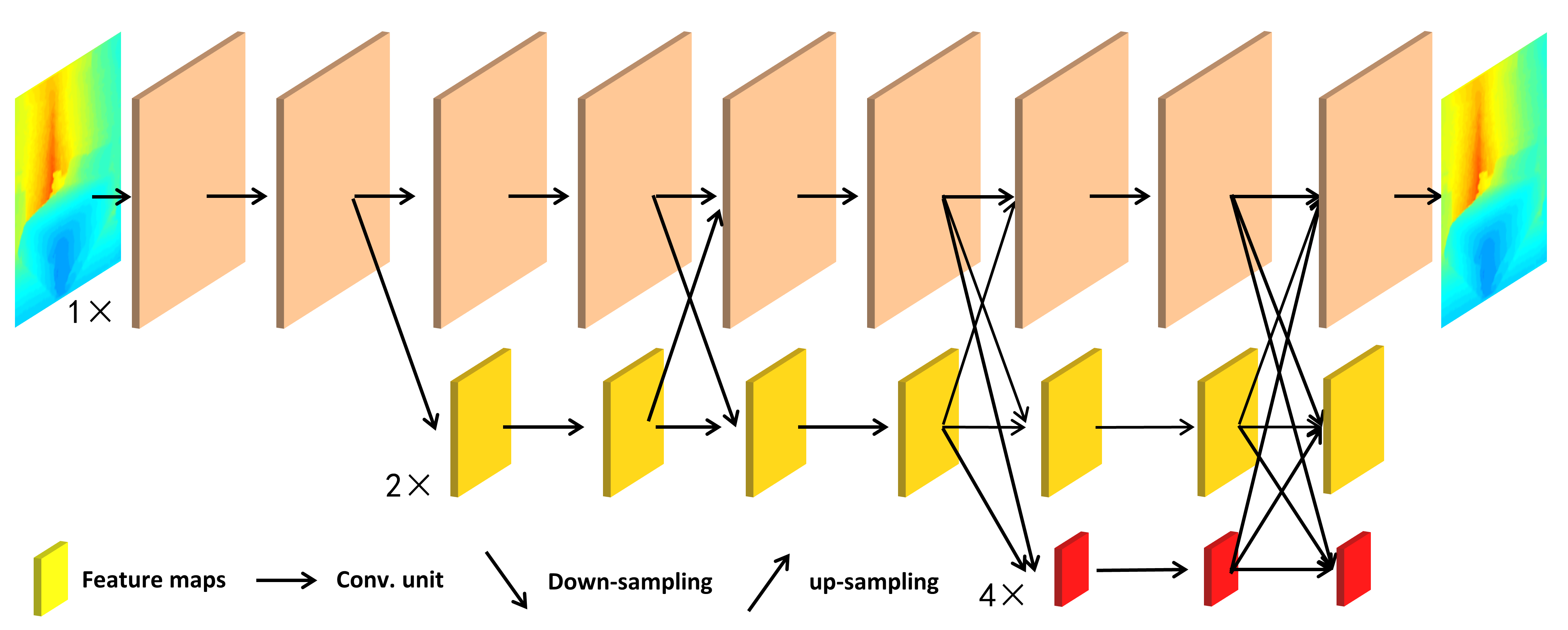}}
\caption{High-Resolution Net architecture.}
\label{HRnet}
\end{figure*}

Numerous existing end-to-end networks are available for the {\bf PFN} backbone, such as U-net\cite{ronneberger2015u}, ResNet\cite{he2016deep}, Hourglass network\cite{newell2016stacked} and deep High-Resolution network(HR-net)\cite{sun2019deep}.
In previous works, most end-to-end networks are directly transfer true-colour image to depth map. 
The backbone of {\bf PFN} demonstrates the preliminary depth estimation, which requires further refinement in the following module. In order to employ a more effective and deeper network, HR-net\cite{sun2019deep} is chosen to be the backbone and shown as Fig.
\ref{HRnet}.

\subsubsection{Spatial Attention Residual Refinement Module}

\begin{figure*}[htbp]
\centerline{\includegraphics[width=0.8\linewidth]{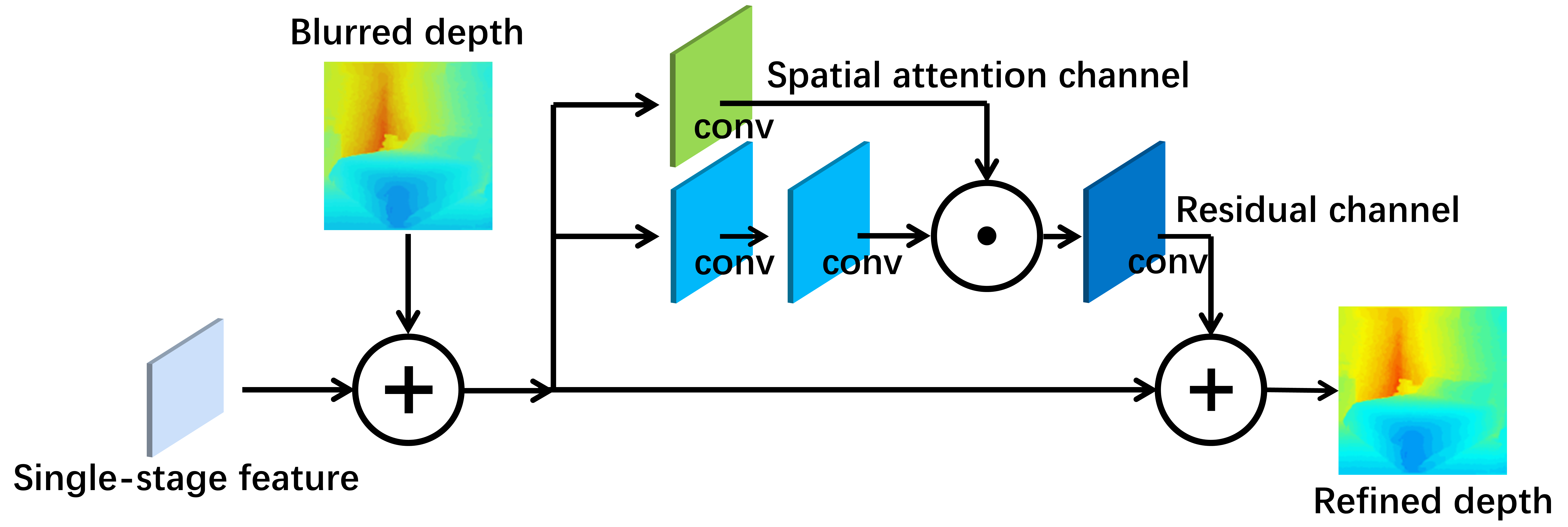}}
\caption{Spatial Attention Residual Refinement Module.}
\label{SARRM}
\end{figure*}

The key idea of the refinement module is to define the intuitive combination of blur depth and the features from different frequency bands.
Unlike the existing fusion methods, we place local residual modules with spatial attention which guide periodic features to refine the blur depth.
To reconstruct depth maps with more accurate details, the Spatial Attention Residual Refinement Module(SARRM) is proposed to replace the direct convolution method. 
Firstly, the Residual Refinement Module provides a path to extract deeper features at minimal computational loss.
Then, the spatial attention mechanism improves information efficiency by enhancing spatial relationships.
The architecture of SARRM is shown in Fig.\ref{SARRM}.

In contrast to the single-layer convolution module, SARRM can be used as multiple skip connections inside the decoding phase, allowing complicated relations between periodic features and the refined prediction.
The spatial attention mechanism is dedicated to extracting the intrinsic relationships between spatial features.
In terms of network complexity, the SARRM is acceptable in various application scenarios.

\subsection{Loss function}

We employ pixel-wise L2 loss, Mean Square Error(MSE), to measure the estimation error and then combine it with L1 loss, Mean Average Error(MAE), to fine-tune the {\bf PFN}. 
The mix loss function is thus shown as follows:

\begin{equation}
\begin{split}
L_{mix}&=L_{L1}+L_{L2}\\
&=\frac{1}{n}\sum_{i=1}^{n}\left | {y_i-\hat{y_i}}\right |+\frac{1}{n}\sum_{i=1}^{n}({y_i-\hat{y_i}})^2
\end{split}
\end{equation}
where $y$ represents the ground-truth depth, $\hat{y}$ represents the predicted depth and $n$ represents its number of data pairs in the dataset.

In the mixed loss function, the L2 loss, which has a higher order than L1 loss, guides the optimization process in a more effective gradient descent direction.
Because the derivative of the first-order function includes a constant term, the L1 loss is regarded as an optimization-related regularization term.

\section{Experiment}

In this section, we first describe the implementation details, the datasets information and benchmark performance of experiments. The {\bf PFN} outperforms other exist monocular depth estimation methods and established a way to combine frequency components.

\subsection{Implementation details}

In our work, The {\bf PFN} is trained by Adam optimizer~\cite{kingma2014adam} with $\beta_{1}=0.99$, $\beta_{2}=0.999$, $\epsilon=10^{-5}$. 
The learning rate of the model optimizer is set to $10^{-4}$ initially and decreases to half every $100$ epochs. 
In the training parameter set, the batch of data is set to $1$.
The PyTorch implement of {\bf PFN} is trained with one RTX 2080 GPU, and this project is proposed online.

The implementation of 2D-FFT is accelerated by the PyTorch-fft package, which can be directly implemented by pixel vectorization in PyTorch.

\subsection{Datasets}

To evaluate the depth estimation task, we apply three benchmark datasets, Make3D, KITTI depth and NYUv2 datasets. 
The details are elaborated below.

\subsubsection{Make3D dataset}
The Make3D dataset\cite{saxena2008make3d}\cite{saxena2005learning} contains 534 outdoor images, 400 images for training and 134 images for testing. 
The dataset has images with a resolution of 2272$\times$1704 and depth maps with 55$\times$305. 
Therefore, we resize the resolution of all the images to 512$\times$512 and depth maps to 512$\times$512.
Same as the pre-processing of previous works, the depth map $C1$, which range from 0$m$ to 80$m$, and $C2$, which range from 0$m$ to 70$m$, is reported by three commonly used evaluation metrics\cite{karsch2014depth}.

\subsubsection{KITTI depth dataset}

The KITTI depth dataset\cite{geiger2013vision} contains 42,382 rectified stereo pairs from 61 scenes, with a typical image size of 1242$\times$375 pixels. 
To adapt to the model input, images and depth maps are resized into 620$\times$188.
The estimated depth map is evaluated by four commonly used evaluation metrics, which is REL, SqRel, RMS, RMS$_{log}$, and three accuracy of different thresholds ($1.25$, $1.25^{2}$, $1.25^{3}$). 

\subsubsection{NYUv2 dataset}

The NYU Depth v2 dataset\cite{silberman2012indoor} contains 464 indoor video scenes captured by a Microsoft Kinect camera.
{\bf GA-CSPN} is trained on all images from 249 training scenes, and tested on 694 image testing scenes.
To adapt to the model input, images are resized into 512$\times$512 and depth maps are resized into 512$\times$512.
Following previous works, the depth map is evaluated by six commonly used evaluation metrics\cite{karsch2014depth}.

\subsection{Evaluation method}

Following previous works, the depth map is evaluated by absolute relative error(REL), root mean square error(RMS), root mean square logarithmic error(RMS$_{log}$)\cite{karsch2014depth} and three accuracy of different thresholds ($1.25$, $1.25^{2}$, $1.25^{3}$). 
The specific formulas of these evaluation methods are shown below.

\begin{itemize}
	\item {Absolute relative error}
	\begin{equation}
	\begin{split}
	REL=\frac{1}{n}\sum \left |\frac{y_{pred}-y_{gt}}{y_{gt}} \right |
	\end{split}
	\end{equation}
	$y_{pred}$ is depth estimated by network and $y_{gt}$ is the ground truth.
	\item {Root mean square error}
	\begin{equation}
	\begin{split}
	RMS=\sqrt{\frac{1}{n}\sum ({y_{pred}-y_{gt}})^{2}}
	\end{split}
	\end{equation}
	\item {Root mean square logarithmic error}
	\begin{equation}
	\begin{split}
	RMS_{log}=\sqrt{\frac{1}{n}\sum ({log(y_{pred})-log(y_{gt}}))^{2}}
	\end{split}
	\end{equation}
	\item {Root mean square logarithmic error}
	\begin{equation}
	\begin{split}
	RMS_{log}=\sqrt{\frac{1}{n}\sum ({log(y_{pred})-log(y_{gt}}))^{2}}
	\end{split}
	\end{equation}
	\item {Accuracy}
	Percentage of depth $y$ where the ratio of estimated and ground truth depth is less than a threshold.
	\begin{equation}
	\begin{split}
	max(\frac{y_{pred}}{y_{gt}},\frac{y_{gt}}{y_{pred}})=\delta< threshold
	\end{split}
	\end{equation}
\end{itemize}

\subsection{Implementation}

We first determine the cut-off frequencies of {\bf PFN} through reasonable ablation experiments.
Then we test DepthTransfer\cite{karsch2014depth}, LRC-Deep3D\cite{xie2016deep3d}, Monodepth\cite{godard2017unsupervised}, {\bf MS-CRF}\cite{xu2017multi}, {\bf DORN}\cite{fu2018deep}, {\bf CSPN}\cite{cheng2018depth}, {\bf GA-CSPN}\cite{lu2021ga} {\bf PFN} and other state-of-the-arts on the standard test sets Make3D\cite{saxena2008make3d}\cite{saxena2005learning}, KITTI depth\cite{geiger2013vision} and NYUv2 datasets\cite{silberman2012indoor}. 
All our experiments were run on the same machine with 32GB of RAM and one RTX 2080Ti GPU.
Each experiment was conducted with an independent model during the training phase.

\subsubsection{Ablation experiments}

In order to select the appropriate cut-off frequency, we compare the accuracy and speed of {\bf PFN} the with multiple cut-off frequencies over the NYUv2 dataset.

The cut-off frequency selection is a significant contributory factor to the improvement of {\bf PFN}. 
The experiments that follow aim to divide the multi-stage information in an appropriate method, thereby allocating valid information reasonably in frequency domains and identifying potential independent information that needs to be interacted with.
In other word, this method extracts features independently from multi-stage frequency bands and avoids the homogeneity of data when the whole image is extracted.

\begin{table*}[htbp]
\caption{Cut-off frequency comparison on NYUv2 dataset}
\label{table}
\centering
{\begin{tabular}{ccccccccc}
\hline
\hline	
F$_{\textit{cut-off}} $  /Hz         & 1            & 5            & 10           & 50           & 100          					& 	REL 		& RMS$_{log}$ & RMS \\	\hline
\multirow{5}{*}{Layer=2} & $\checkmark$ &              &              &              &             					&   0.2520 	& 0.0980      	& 0.9610		 \\
                         &              & $\checkmark$ &              &              &              							&   0.2370	& 0.0950        & 0.9180	 \\
                         &              &              & $\checkmark$ &              &              							&   0.2100	& 0.0890     	& 0.8320   		\\
                         &              &              &              & $\checkmark$ &              							&   0.1960 	& 0.0860 	& 0.7880		 \\
                         &              &              &              &              & $\checkmark$ 							&   0.1820 	& 0.0830      	& 0.7450     	\\	\hline
\multirow{4}{*}{Layer=3} & $\checkmark$ &              & $\checkmark$ &         &          				&   0.1890   	& 0.0870 	& 0.7630		\\
                         		&              & $\checkmark$ & $\checkmark$ &              &              				&   0.1765   	& 0.0802  	& 0.7300 		\\
                         		&              &              & $\checkmark$ & $\checkmark$ &              				&   0.1399	& 0.0620     	& 0.6164 		\\
                         		&              &              & $\checkmark$ &              & $\checkmark$ 				&   0.1220 	& 0.0530      	& 0.5610 	\\		\hline
Layer=4                  & $\checkmark$ &              & $\checkmark$ &              & $\checkmark$ 	  		&   0.1180  	& 0.0500        & 0.4220    	\\	\hline
\multirow{5}{*}{Layer=5} & $\checkmark$ & $\checkmark$ & $\checkmark$ & $\checkmark$ &		&   0.1200      & 0.0500    	& 0.4240           \\	
                         & $\checkmark$ & $\checkmark$ & $\checkmark$ &              & $\checkmark$ 		&   0.1150   	&  0.0490       & 0.4160  \\
                         & $\checkmark$ & $\checkmark$ &              & $\checkmark$ & $\checkmark$ 		&   0.1100   	& 0.0490        & 0.4080   \\
                         & $\checkmark$ &              & $\checkmark$ & $\checkmark$ & $\checkmark$ 		&   0.1030      & 0.0470       	& 0.3970   \\
                         &              & $\checkmark$ & $\checkmark$ & $\checkmark$ & $\checkmark$ 		&   0.1010 	& 0.0460	& 0.3920      \\	\hline
Layer=6                  & $\checkmark$ & $\checkmark$ & $\checkmark$ & $\checkmark$ & $\checkmark$ &    0.1000  & 0.0470    & 0.3950        \\       
\hline
\hline	
\end{tabular}}
\label{tab4layeracc}
\end{table*}

Therefore, ablations of the cut-off frequencies in {\bf PFN} on the NYUv2 dataset are shown in Table.\ref{tab4layeracc}.
The experiment result in Fig.\ref{Acc4freqprob} shows that most depth information is preserved in the frequency band between 5Hz and 30Hz.
Thus, in the ablation experiment, dense cut-off frequencies are applied in the low-frequency part, while sparse cut-off frequencies are used in the high-frequency part.
With the multi-layer modeling, the minimum REL of {\bf PFN} is achieved at 0.1000, and the RMS is achieved at 0.0460.

Results on stronger backbones, by replacing U-net with HR-net\cite{sun2019deep}, are presented in Table.\ref{tab4layerspeed}.
As seen in Table.\ref{tab4layerspeed}, the complex network structure of {\bf PFN} results in the whole algorithm failing to operate in real-time on various sizes of input images.
The increasing number of layers is found to be particularly effective in improving accuracy, and the five-layer pyramid network 
maintains a trade-off between accuracy and running time.
Therefore, the cut-off frequencies of {\bf PFN} are set as 5Hz, 10Hz, 50Hz and 100Hz.

\begin{table*}[htbp]
\caption{Cut-off frequency comparison for PFN on NYUv2 dataset}
\label{table}
\centering
{\begin{tabular}{crrrr}
\hline
\hline	
 Layer 	&320$\times$240	&800$\times$600	&1024$\times$768\\  \hline
 2  & 64.025$ms$ 		& 	409.590$ms$       	& 	1265.915$ms$  \\
  3  & 95.823$ms$		& 	497.908$ms$      	& 	1382.964$ms$ \\ 
  4  &128.739$ms$ 		&	595.256$ms$	&	1504.640$ms$ \\
  5  &156.018$ms$ 		&	658.708$ms$	&	1602.502$ms$ \\
 6  &181.680$ms$ 		&	734.044$ms$	&	1688.506$ms$ \\
\hline
\hline	
\end{tabular}}
\label{tab4layerspeed}
\end{table*}

\subsubsection{Quantitative comparison}

In this section, we compare the complexity and the accuracy between state-of-the-art strategies quantitatively.

\begin{table}[htb]
\caption{Comparison of parameter number and running time on NYUv2 dataset}
\label{table}
\centering
{\begin{tabular}{lcrrrl}
\hline
\hline
Method & \#Params & Running time\\ \hline

 CSPN\cite{cheng2018depth}   & 17.1M 	&	139.698$ms$	 \\
 GA-CSPN\cite{lu2021ga}      & 21.3M 	&	171.853$ms$	\\
 DORN\cite{fu2018deep}	  & 51.1M 	&	276.255$ms$	\\
 BTS\cite{lee2019big}        & 47.0M 	&	860.589$ms$	\\
 PFN      				  & 57.2M 	&	561.400$ms$\\
\hline
\hline
\end{tabular}}
\label{tab4ParamTime}
\end{table}

Table.~\ref{tab4ParamTime} demonstrates parameter number and running time of state-of-the-art models on NYUv2 dataset.
It can be observed from Table.~\ref{tab4ParamTime} that the parameter number of {\bf PFN} is a little more than other state-of-the-art methods. 
Compared to the BTS\cite{lee2019big}, the parameter number of {\bf PFN} is only increased by 20.2\%, but the running time of {\bf PFN} is decreased by 34.8\%.

Table.\ref{tab4complexMake} shows quantitative comparison results on the Make3D dataset over state-of-the-art strategies.
It displays the REL, RMS$_{log}$ and RMS values, and these values represent averages over all the images on the Make3D dataset.
Recall that lower REL, RMS$_{log}$ and RMS values reflect better performance. 
It can be observed that {\bf PFN}, with the proper frequency selection, yields the best performance in most evaluation methods.
With the proposed frequency-based method, the minimum REL of {\bf PFN} is achieved at 0.1480, and the RMS is achieved at 3.8180 in the C1 metric.
All the learning-based feature extraction methods systematically give better results than traditional methods.

\begin{table*}[htbp]
\caption{Comparison results on Make3D dataset between state-of-the-art strategies}
\label{table}
\centering
{\begin{tabular}{lcccccc}
\hline	
\hline	
\multirow{2}{*}{Method} & \multicolumn{3}{c}{C1}	& \multicolumn{3}{c}{C2}	\\

			&	REL		&	RMS$_{log}$&	RMS		&	REL		&	RMS$_{log}$	&RMS	\\
\hline
Make3D\cite{saxena2008make3d}  		&	-		&	-		&	-		&	0.3700 	&	0.1870 	&	-	\\
Liu et al.\cite{liu2010single} 	&	-		&	-		&	-		&	0.3790 	&	0.1480 	&	-	\\
DepthTransfer\cite{karsch2014depth} 	&	0.3550 	&	0.1270 	&	9.2000 	&	0.3610 	&	0.1480 	&	15.1000 	\\
Liu et al.\cite{liu2014discrete} 	&	0.3350 	&	0.1370 	&	9.4900 	&	0.3380 	&	0.1340 	&	12.6000 	\\
Li et al.\cite{li2015depth} 		&	0.2780 	&	0.0920 	&	7.1200 	&	0.2790 	&	0.1020 	&	10.2700 	\\
Liu et al.\cite{liu2015learning}		&	0.2870 	&	0.1090 	&	7.3600 	&	0.2870 	&	0.1220 	&	14.0900 	\\
Roy et al.	\cite{roy2016monocular}	&	-		&	-		&	-		&	0.2600 	&	0.1190 	&	12.4000 	\\
Laina et al.\cite{laina2016deeper} 	&	0.1760 	&	0.0720 	&	4.4600 	&	-		&	-		&	-	\\
LRC-Deep3D\cite{xie2016deep3d} 	&	1.0000 	&	2.5270 	&	19.1100 	&	-		&	-		&	-	\\
Monodepth\cite{godard2017unsupervised} 			&	0.4430 	&	0.1560 	&	11.5130 	&	-		&	-		&	-	\\
Kuznietsov et al.\cite{kuznietsov2017semi} &	0.4210 	&	0.1900 	&	8.2400 	&	-		&	-		&	-	\\
MS-CRF\cite{xu2017multi}		&	0.1840 	&	0.0650 	&	4.3800 	&	0.1980 	&			&	8.5600 	\\
DORN\cite{fu2018deep} 			&\color{blue}{0.1570} 	&0.0620 				&\color{blue}{3.9700} 	& 0.1620 	&	\color{blue}{0.0670} 	&\color{red}{7.3200} 	\\	
CSPN\cite{cheng2018depth}		&	0.2360 			&	0.0800 				&	6.0480 			&	0.2370 	&	0.1090 	&	12.0110 	\\	
GA-CSPN\cite{lu2021ga}			&	0.1540 			&	\color{blue}{0.0580} 	&	3.7880		&	\color{blue}{0.1470} 	&	\color{red}{0.0580} 	&	7.8600 	\\
PFN						&	\color{red}{0.1480} &	\color{red}{0.0500} 	&	\color{red}{3.8180} 	&	\color{red}{0.1450} 	&	\color{red}{0.0580} 	&	\color{blue}{7.8400} 	\\

\hline
\hline
\multicolumn{6}{p{240pt}}{Red numbers mark the best score.}\\
\multicolumn{6}{p{240pt}}{Blue numbers mark the second best score.}\\
\end{tabular}}
\label{tab4complexMake}
\end{table*}

As quantitatively shown in Table.\ref{tab4complexKITTI}, the proposed end-to-end approach significantly outperforms both the existing fully-supervised and semi-supervised methods on the KITTI depth dataset.
On this dataset, the proposed method results in lower errors and higher accuracy compared to the existing fully-supervised methods. 
Without using extra data argumentation methods, the proposed method achieves 0.0690 REL on test, which is over 4\% relative improvement over the previous best method\cite{lu2021ga}, and is also better than the concurrent work of Monodepth2's 0.1060\cite{godard2019digging}.
With the proposed frequency-based method, the minimum REL of {\bf PFN} is achieved at 0.1480, and the RMS is achieved at 3.8180 in C1 metric.

\begin{table*}[htbp]
\caption{Comparison results on KITTI depth dataset between state-of-the-art strategies}
\label{table}
\centering
{\begin{tabular}{lccccccccc}
\hline
\hline
Method 		&	Method	&	REL	&	SqRel	&	RMS	&	RMS$_{log}$	&	$\delta < 1.25$	&	$\delta < 1.25^2$	&	$\delta < 1.25^3$	\\
\hline
GeoNet\cite{yin2018geonet}	&	S 	&	0.1490	&	1.0600	&	5.5670	&	0.2260	&	0.7960	&	0.9350	&	0.9750	\\
DDVO\cite{wang2018learning}	&	S 	&	0.1510	&	1.2570	&	5.5830	&	0.2280	&	0.8100	&	0.9360	&	0.9740	\\
DF-Net\cite{zou2018df}	&	S 	&	0.1500	&	1.1240	&	5.5070	&	0.2230	&	0.8060	&	0.9330	&	0.9730	\\
Monodepth\cite{godard2017unsupervised} 	&	S 	& 	0.1330 	&	1.1420 	&	5.5330 	&	0.2300 	&	0.8300	& 	0.9360 	&	0.9700\\
Monodepth2\cite{godard2019digging}	&	S 	& 	0.1060	& 	0.8060 	&	4.6300 	&	0.1930 	&	0.8760 	&	0.9580 	&	0.9800\\
\hline
Eigen et al.\cite{eigen2015predicting}	&	D 	&	0.2030 	&	1.5480 	&	6.3070 	&	0.2820 	&	0.7020 	&	0.8900 	&	0.8900 	\\
Liu et al.\cite{liu2015learning}	&	D 	&	0.2010 	&	1.5840 	&	6.4710 	&	0.2730 	&	0.6800 	&	0.8980 	&	0.9670 	\\
AdaDepth\cite{kundu2018adadepth}		&	D 	&	0.1670 	&	1.2570 	&	5.5780 	&	0.2370 	&	0.7710 	&	0.9220 	&	0.9710 	\\
Kuznietsov et al.\cite{kuznietsov2017semi}	&	D 	&	0.1130 	&	0.7410 	&	4.6210 	&	0.1890 	&	0.8620 	&	0.9600 	&	0.9860 	\\
DVSO\cite{yang2018deep}	&	D 	&	0.0970 	&	0.7340 	&	4.4420 	&	0.1870 	&	0.8880 	&	0.9580 	&	0.9800 	\\
SVSM FT\cite{luo2018single}	&	D 	&	0.0940 	&	0.6260 	&	4.2520 	&	0.1770 	&	0.8910 	&	0.9650 	&	0.9840 	\\
Guo	et al.\cite{guo2018learning} &	D 	&	0.0960 	&	0.6410 	&	4.0950 	&	0.1680 	&	0.8920 	&	0.9670 	&	0.9860 	\\
DORN\cite{fu2018deep} 	&	D 	&	\color{blue}{0.0720} 	&	\color{red}{0.3070} 	&	2.7270 	&	0.1200 	&	0.9320 	&	0.9840 	&	0.9940 	\\
CSPN\cite{cheng2018depth}		&	D 	&	0.1750 	&	1.3620 	&	5.9770 	&	0.2450 	&	0.7600 	&	0.9140 	&	0.9690	\\
GA-CSPN\cite{lu2021ga}	      	&	D 	&	\color{blue}{0.0720} 	&	\color{blue}{0.3250} 	&	\color{blue}{2.7020} 	&	\color{blue}{0.1160} 	&	\color{blue}{0.9510} 	&	\color{red}{0.9890} 	&	\color{red}{0.9960} \\
PFN	      	&	D 	&	\color{red}{0.0690} 	&	\color{red}{0.3020} 	&	\color{red}{2.6520} 	&	\color{red}{0.1120} 	&	\color{red}{0.9530} 	&	\color{red}{0.9890} 	&	\color{blue}{0.9950} \\

\hline
\hline
\multicolumn{6}{p{200pt}}{D - Depth supervision}\\
\multicolumn{6}{p{200pt}}{S - Self-supervised stereo supervision}\\
\end{tabular}}
\label{tab4complexKITTI}
\end{table*}

Quantitative results on the NYUv2 dataset are shown in Table.\ref{tab4complexNYU}, and qualitative results are shown in Fig.\ref{nyuq}.
Without any tricks for data argumentation on the NYUv2 dataset,{\bf PFN} gives a marginal improvement of 1.85\% RMS over the previous best method, AdaBins\cite{bhat2020adabins}, on the test set. 
In the NYUv2 dataset, the proposed method has reached the best in six evaluation methods.
Conclusively, the proposed {\bf PFN} outperforms all existing methods in all evaluation methods on the NYUv2 dataset.

\begin{table*}[htbp]
\caption{Comparison results on NYUv2 dataset between state-of-the-art strategies}
\label{table}
\centering
{\begin{tabular}{lccccccl}
\hline
\hline
Method 		&	REL		&	RMS$_{log}$&	RMS 		&$\delta < 1.25$&$\delta < 1.25^2$&$\delta < 1.25^3$	\\
\hline
Make3D\cite{saxena2008make3d} 		&	0.3490 	&	-		&	1.2140 	&	0.4470 	&	0.7450 	&	0.8970 	\\
DepthTransfer\cite{karsch2014depth}	&	0.3500 	&	0.1310 	&	1.2000 	&	-	&	-	&	-	\\
Liu et al.\cite{liu2014discrete}			&	0.3350 	&	0.1270 	&	1.0600 	&	-	&	-	&	-	\\
Ladicky et al.\cite{ladicky2014pulling} 	&	-	&	-	&	-	&	0.5420 	&	0.8290 	&	0.9410 	\\
Li et al.\cite{li2015depth} 				&	0.2320 	&	0.0940 	&	0.8210 	&	0.6210 	&	0.8860 	&	0.9680 	\\
Wang et al.\cite{wang2015towards} 		&	0.2200 	&	-	&	0.8240 	&	0.6050 	&	0.8900 	&	0.9700 	\\
Roy et al.\cite{roy2016monocular} 		&	0.1870 	&	-	&	0.7440 	&	-	&	-	&	-	\\
Liu et al. \cite{liu2015learning}			&	0.2130 	&	0.0870 	&	0.7590 	&	0.6500 	&	0.9060 	&	0.9760 	\\
Eigen et al.\cite{eigen2015predicting} 	&	0.1580 	&	-	&	0.6410 	&	0.7690 	&	0.9500 	&	0.9880 	\\
Chakrabarti et al.\cite{chakrabarti2016depth} 	&	0.1490 	&	-	&	0.6200 	&	0.8060 	&	0.9580 	&	0.9870 	\\
Laina et al.\cite{laina2016deeper} 		&	0.1270 	&	0.0550 	&	0.5730 	&	0.8110 	&	0.9530 	&	0.9880 	\\
Li et al.\cite{li2017two}					&	0.1430 	&	0.0630 	&	0.6350 	&	0.7880 	&	0.9580 	&	0.9910 	\\
MS-CRF \cite{xu2017multi}		&	0.1210	&	0.0520 	&	0.5860 	&	0.8110 	&	0.9540 	&	0.9870 	\\
DORN\cite{fu2018deep}				&	0.1150 	&	0.0510 	&	0.5090 	&	0.8280 	&0.9650		&	0.9900 	\\
SharpNet\cite{ramamonjisoa2019sharpnet} &	0.1390	&	\color{blue}{0.0470} 	&	0.5020 	&	0.8360 	&	0.9660 	&	0.9900 	\\
BTS\cite{lee2019big} 					&	\color{blue}{0.1100}	&	\color{blue}{0.0470} 	&	\color{blue}{0.3950}	&	0.8450 	&	0.9730 	&	0.9920 	\\
AdaBins\cite{bhat2020adabins}			&	\color{blue}{0.1100}	&	0.0480	&	\color{red}{0.3920} 	&	\color{blue}{0.8850}			&	0.9780 		&	\color{red}{0.9940}	\\
CSPN\cite{cheng2018depth}			&	0.1950 			&	0.0640 				&	0.6970 	&	0.7620 	&	0.9250 				&	0.9720	\\
GA-CSPN\cite{lu2021ga}	      						&	0.1090 			&	0.0450 				&	0.4030 	&	0.8710 	&	\color{blue}{0.9790} 	&	\color{blue}{0.9930} 	\\
PFN								&	\color{red}{0.1010} &	\color{red}{0.0430}	&	\color{red}{0.3920} 	&	\color{red}{0.8870} &	\color{red}{0.9810} 	&	\color{red}{0.9940} 	\\

\hline
\hline
\multicolumn{6}{p{240pt}}{Red numbers mark the best score.}\\
\multicolumn{6}{p{240pt}}{Blue numbers mark the second best score.}\\
\end{tabular}}
\label{tab4complexNYU}
\end{table*}

Above experiments demonstrate that the five-layer {\bf PFN} yields better performance than other existing model-based methods.
The improvement of accuracy is expected because of the increased capacity of the network.
However, it also results in longer training times. 
Hence, we resort to using the five-layer pyramid as the standard architecture in all our experiments.
These experiments confirmed that the {\bf PFN} is better than the state-of-the-art for most metrics on Make3D\cite{saxena2008make3d}\cite{saxena2005learning}, KITTI depth\cite{geiger2013vision} and NYUv2 datasets\cite{silberman2012indoor} datasets.

\subsubsection{Qualitative comparison}

In qualitative comparison, we test DepthTransfer\cite{karsch2014depth}, Monodepth2\cite{godard2019digging}, {\bf DORN}\cite{fu2018deep}, {\bf CSPN}\cite{cheng2018depth} and the proposed {\bf PFN} on the standard test sets Make3D\cite{saxena2008make3d}\cite{saxena2005learning}, KITTI depth\cite{geiger2013vision} and NYUv2 datasets\cite{silberman2012indoor}. 

\begin{figure*}[ht]
\centerline{\includegraphics[width=0.8\linewidth]{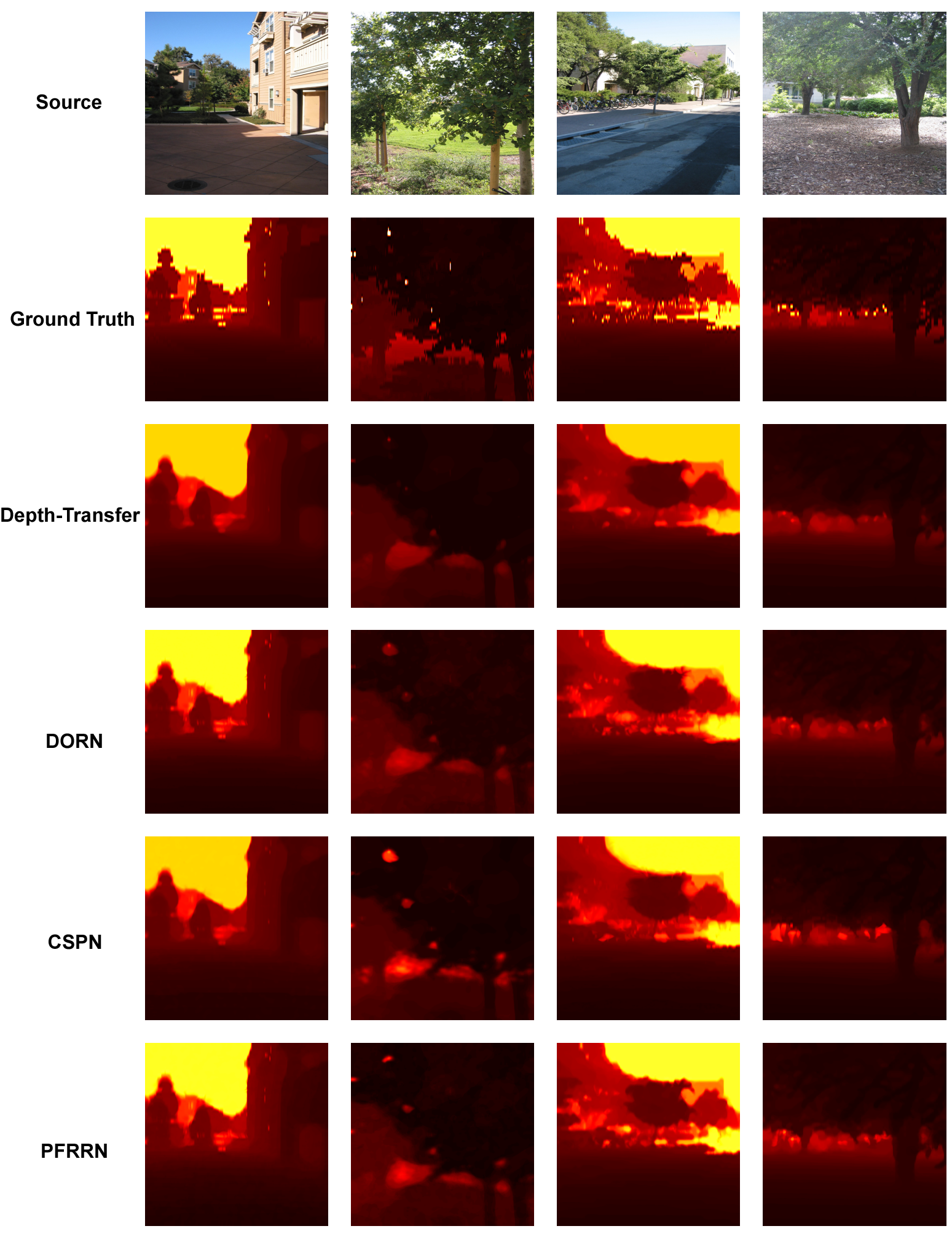}}
\caption{Qualitative comparison on Make3D}
\label{make3dq}
\end{figure*}

Results of depth estimation on the Make3D\cite{saxena2005learning} dataset are shown in Fig.\ref{make3dq}.
Compared with NYUv2 and KITTI depth dataset, Make3D contains a relatively small number of samples.
Therefore, the inferior visual performance of most deep-learning methods on Make3D due to over-fitting, such as the mismatch of colour and object.
While {\bf CSPN} and {\bf DORN} mostly restore the depth from input, many of their results are shown as blurry and do not maintain the degree of sharpness as seen in the ground truth. 
Depth-Transfer even fails to reconstruct the proper edge between the object and background in the image by the style-transfer method.
The {\bf PFN} clearly reconstructs the most detailed depth map while properly smoothing the low-resolution ground truth.

\begin{figure*}[ht]
\centerline{\includegraphics[width=\linewidth]{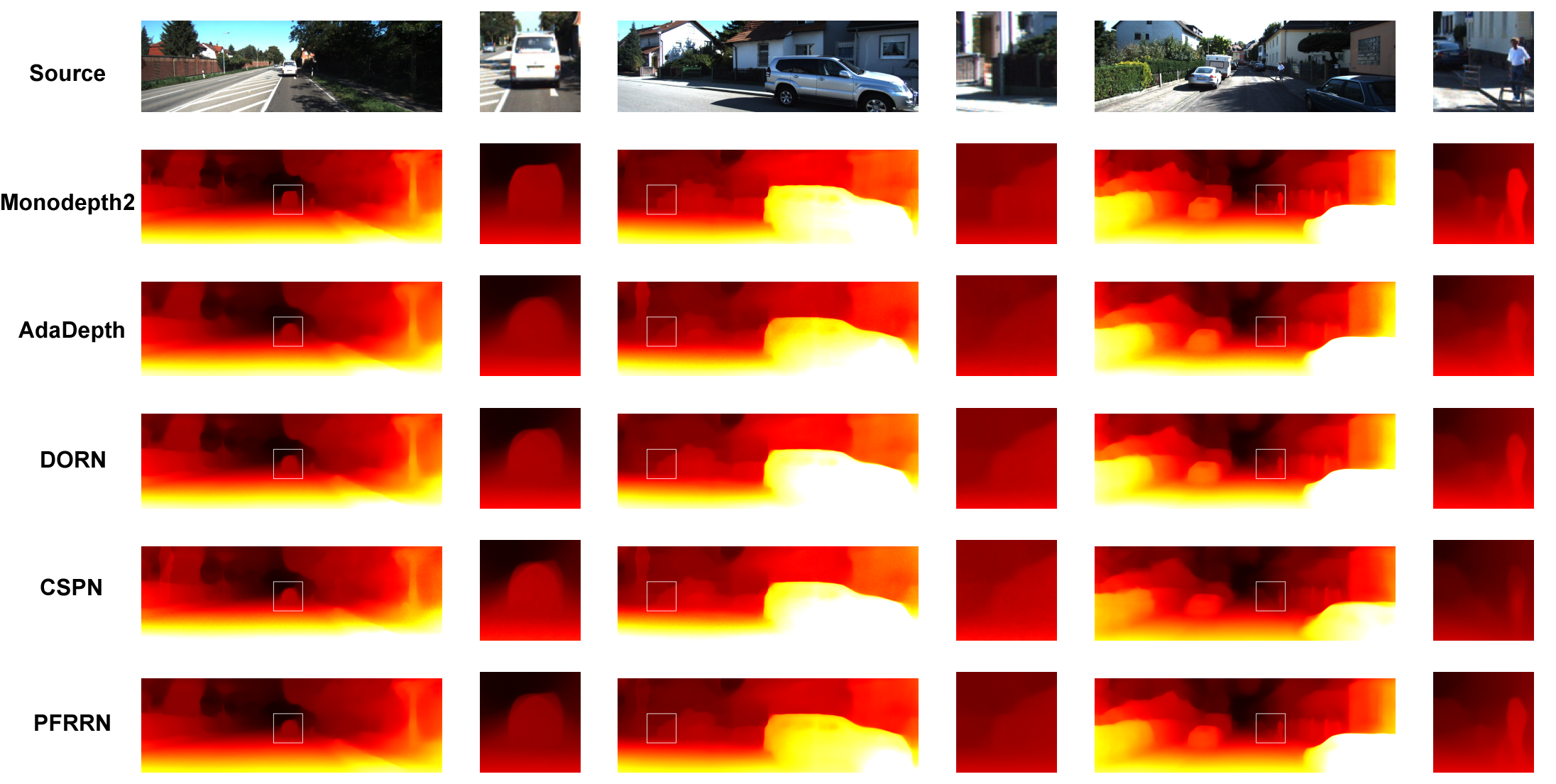}}
\caption{Qualitative comparison on KITTI depth dataset}
\label{kittiq}
\end{figure*}

Reconstruction cases and their corresponding zoomed details of depth estimation on the KITTI depth dataset are shown in Fig.\ref{kittiq}.
In zoomed depth maps, the depth map of {\bf PFN} shows much more precise object boundaries.
The car shown in the zoomed rectangle is reconstructed accurately by the proposed {\bf PFN} algorithm compared to the other algorithms. 
In addition, {\bf PFN} has a good visual performance for the zoomed car, but shows a slight visible ringing artifact on object boundaries.
It can also be observed that there is a more accurate pedestrian reconstruction in the lane frame and a better reconstruction of the bush in
the garden.

\begin{figure*}[ht]
\centerline{\includegraphics[width=0.8\linewidth]{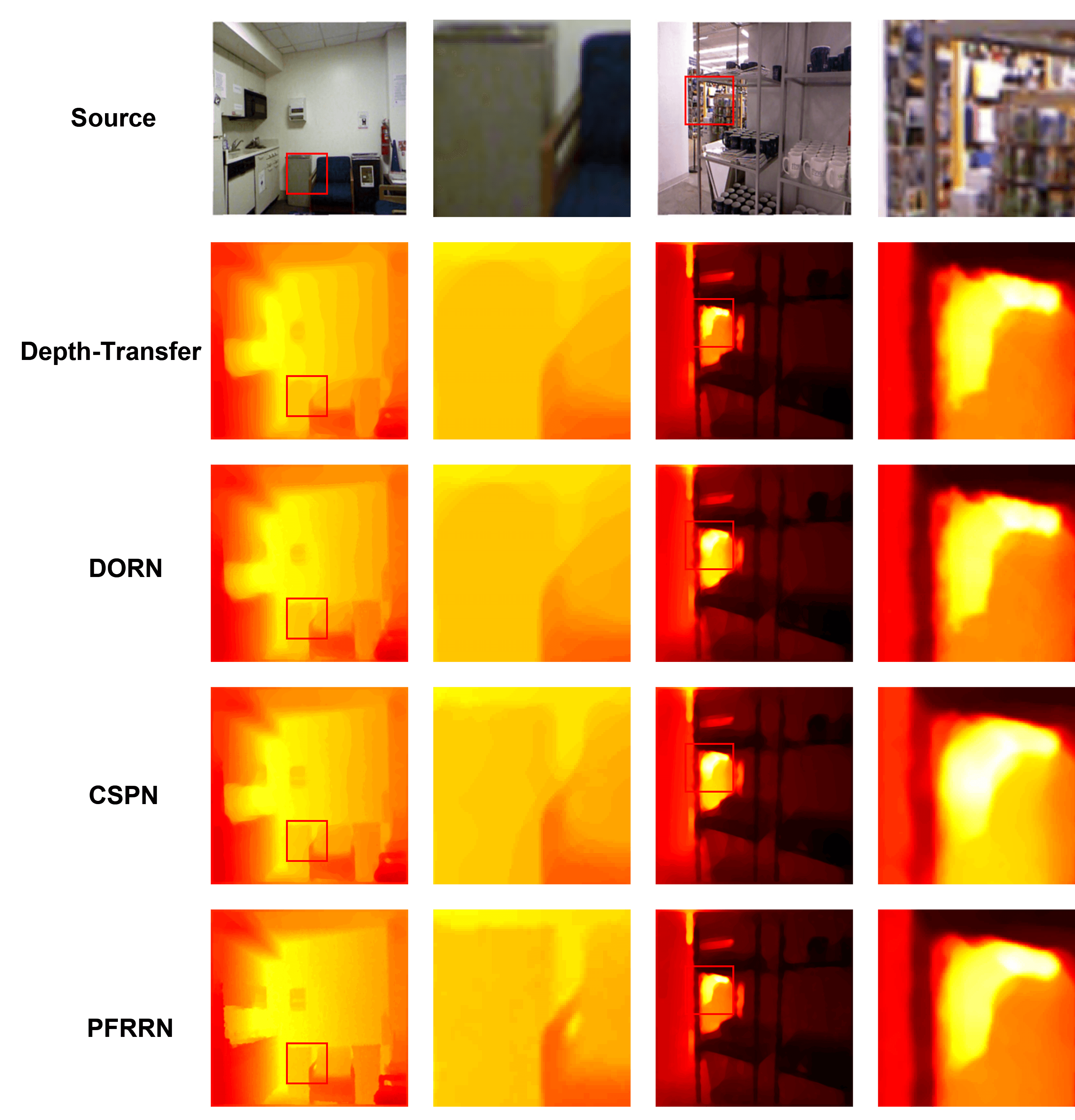}}
\caption{Qualitative comparison on NYUv2 dataset}
\label{nyuq}
\end{figure*}

On the NYUv2 dataset, Fig.\ref{nyuq} shows the qualitative depths on the left column and their corresponding zoomed details on the right column.
Due to the slight errors between the estimated depth maps, we provide the zoomed depth maps to further observe the quality of the proposed methods.
It can be seen that local depths which have large gradients, such as the hole in the cup shelf, are hard to reconstruct.
Our {\bf PFN} not only reconstructs the proper depth of the area, but also reconstructs a clear boundary of the bookshelf.
The proposed approach is implemented with the backbone of HR-net\cite{sun2019deep} which establishes feature transfer paths between different layers instead of a single path. 
{\bf PFN}'s superiority in depth quality results from effective frequency feature fusion from multi-layer frequency division.

\subsection{Robustness analysis}

The previous section demonstrated the performance of existing model-based methods on widely-used datasets.
In this section, we evaluate the performance of {\bf PFN} on real-world indoor datasets with noisy environments.
Specifically, we chose Gaussian noises for experiments.
Same as previous experiments, images are resized into 512$\times$512 and depth maps are resized into 512$\times$512 in robustness experiments.

\subsubsection{Gaussian noise}

Gaussian noise is statistical noise having a probability density function equal to that of the normal distribution, which is also known as the Gaussian distribution.
The probability density function $p$ of a Gaussian random variable $z$ is given by:
\begin{equation}
p_{G}(z)={\frac  {1}{\sigma {\sqrt  {2\pi }}}}e^{{-{\frac  {(z-\mu )^{2}}{2\sigma ^{2}}}}}
\end{equation}
where $z$ represents the pixel value, $\mu$ represents the mean value and $\sigma$ represents its standard deviation.

\begin{figure*}[htbp]
\centerline{\includegraphics[width=1\linewidth]{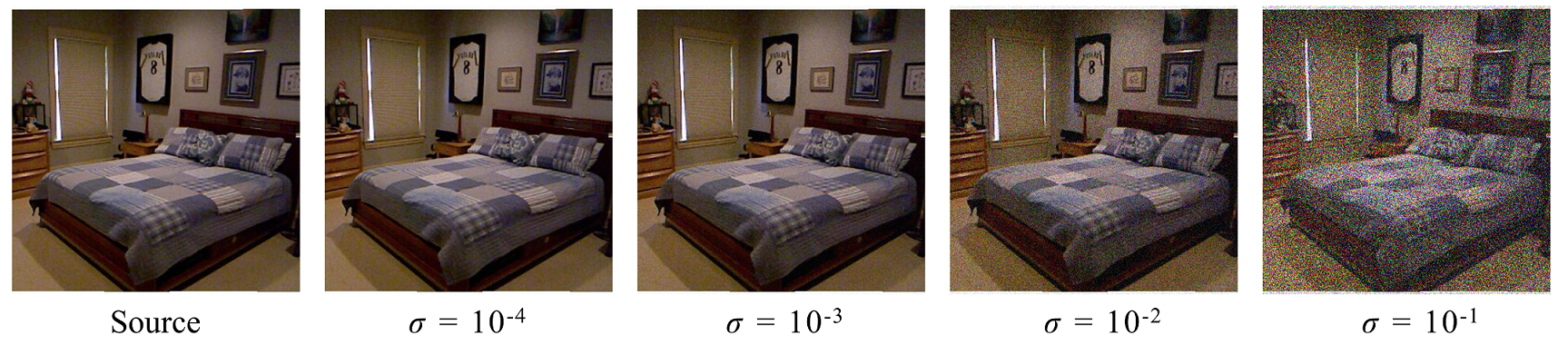}}
\caption{Gaussian noise additive examples on NYUv2 dataset}
\label{fig:noise}
\end{figure*}

For pre-processing, the original images are added with random Gaussian noise and images under different noise variances are shown in Fig.\ref{fig:noise}. $\sigma$ is the variance of Gaussian noise.
From the Fig.\ref{fig:noise}, the slight noise, whose variance $\leq10^{-3}$, are unperceivable for human vision.

\subsubsection{Robustness experiment}

To validate the robustness of existing deep-learning methods, we evaluate the performance of {\bf CSPN}
\cite{cheng2018depth}, {\bf GA-CSPN}\cite{lu2021ga} and {\bf PFN} on the Gaussian noise additive dataset which is modified from the NYUv2 dataset.

\begin{figure*}[htbp]
\centerline{\includegraphics[width=1\linewidth]{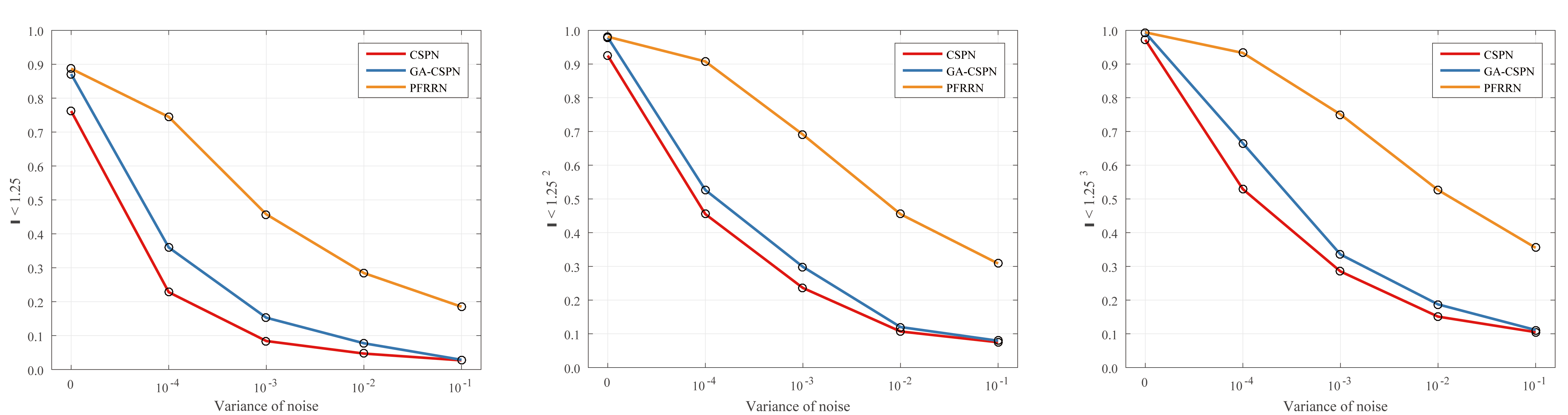}}
\caption{Gaussian noise additive model results on NYUv2 dataset}
\label{fig:AccGau}
\end{figure*}

As shown in Fig.\ref{fig:AccGau}, the NYUv2 dataset with a gradual increase of Gaussian noise is evaluated by three accuracy metrics with different thresholds.
Though the slight noise is unperceivable for human vision, it significantly reduces the accuracy of depth estimation in most works.
Compared with {\bf CSPN} and {\bf GA-CSPN} under Gaussian noise circumstances, the accuracy of {\bf PFN} decreases much less.

As shown above, the proposed method is a robust end-to-end depth estimation approach that surpasses traditional fully-supervised methods.
Because the pyramid frequency module can separate most aperiodic features from input images, the depth prediction is capable of extracting more periodic and valid features. 
Although the accuracy of the {\bf PFN} decreases in the noisy dataset, it has achieved significant progress compared to previous methods.

\section{Conclusion}

We have presented a Pyramid Frequency Network(PFN), a frequency-based network to extract features from multiple frequency bands and find potential independent relations that need to be interacted with.
To reconstruct depth maps with more accurate details, a Spatial Attention Residual Refinement Module(SARRM) is proposed to refine the blur depth.
The proposed {\bf PFN} is tested on three benchmark datasets, Make3D, KITTI depth and NYUv2 datasets and outperforms the state-of-the-art methods.
With extra robustness testing on the noisy NYUv2 dataset, we have confirmed that {\bf PFN} for monocular depth estimation in common scenes is more reliable than depth estimations obtained with state-of-the-art deep-learning methods.

\section*{Acknowledgment}

This work is supported by the National Natural Science Foundation of China(grant no. 62173160 and 61573168).

\bibliographystyle{spiejour}   
\bibliography{reference}   


\vspace{2ex}\noindent\textbf{Zhengyang Lu} received the M.Sc. degree in electrical and electronic engineering from University of Surrey, Guilford, united kingdom in 2018. Currently, he is a phd candidate with Jiangnan University. His current research interests include depth estimation, photometric stereo, and SLAM.

\vspace{2ex}\noindent\textbf{Ying Chen} received the Ph.D. degree in Control Science and Engineering from Xi'an Jiaotong University in 2005. 
She is a professor in the Department of Information Technology at Jiangnan University. Her research activities are focused on computer vision, pattern recognition and information fusion.

\listoffigures
\listoftables

\end{spacing}
\end{document}